\pdfoutput=1

\documentclass[11pt]{article}

\usepackage[final]{acl}

\usepackage{times}
\usepackage{latexsym}

\usepackage[T1]{fontenc}

\usepackage[utf8]{inputenc}

\usepackage{microtype}

\usepackage{inconsolata}

\usepackage{graphicx}
\usepackage{subfigure}
\usepackage{booktabs}
\usepackage{adjustbox}
\usepackage{multirow}
\usepackage{hyperref}

\title{Blending LLMs into Cascaded Speech Translation: KIT's Offline Speech Translation System for IWSLT 2024}

\author{
 \textbf{Sai Koneru},
 \textbf{Thai-Binh Nguyen},
 \textbf{Ngoc-Quan Pham},
 \textbf{Danni Liu},
 \textbf{Zhaolin Li},
\\
 \textbf{Alexander Waibel},
 \textbf{Jan Niehues}
\\
\\
 Karlsruhe Institute of Technology
\\
\\
   \href{mailto:email@domain}{firstname.lastname@kit.edu}
}

\begin{document}
\maketitle
\begin{abstract}

Large Language Models (LLMs) are currently under exploration for various tasks, including Automatic Speech Recognition (ASR), Machine Translation (MT), and even End-to-End Speech Translation (ST). In this paper, we present KIT's offline submission in the constrained + LLM track by incorporating recently proposed techniques that can be added to any cascaded speech translation. Specifically, we integrate Mistral-7B\footnote{mistralai/Mistral-7B-Instruct-v0.1} into our system to enhance it in two ways. Firstly, we refine the ASR outputs by utilizing the N-best lists generated by our system and fine-tuning the LLM to predict the transcript accurately. Secondly, we refine the MT outputs at the document level by fine-tuning the LLM, leveraging both ASR and MT predictions to improve translation quality. We find that integrating the LLM into the ASR and MT systems results in an absolute improvement of $0.3\%$ in Word Error Rate and $0.65\%$ in COMET for tst2019 test set. In challenging test sets with overlapping speakers and background noise, we find that integrating LLM is not beneficial due to poor ASR performance. Here, we use ASR with chunked long-form decoding to improve context usage that may be unavailable when transcribing with Voice Activity Detection segmentation alone.

\end{abstract}
\section{Introduction}

This paper provides an overview of Karlsruhe Institute of Technology's speech translation (ST) system developed for the offline track of IWSLT 2024. We participated in the constrained plus large language models (LLMs) condition, focusing on the translation direction from English to German. Under this condition, LLMs with parameters of around 7 billion are allowed, and they have proven effective in many NLP tasks. One of the interesting aspects of this condition is how one can effectively integrate them into ST systems.

In recent years, there has been a significant interest in developing several open-sourced and medium-scale LLMs \citep{touvron2023llama,jiang2023mistral}. The adaptability of LLMs to diverse tasks, using techniques such as In-Context-Learning \cite{brown2020language} or Parameter-efficient fine-tuning with 4-bit quantization \cite{hu2021lora,dettmers2024qlora}, enables their exploitation even with limited resources.

With these recent advancements, exploiting LLMs for ST shows great promise and offers several potential benefits. For instance, one common challenge in Automatic Speech Recognition (ASR) is dealing with input noise, which can often render it difficult to comprehend the speaker's words. However, LLMs, trained on vast amounts of data, may excel at predicting words compared to decoders trained solely during ASR. Moreover, LLMs possess a richer vocabulary and understanding of complex terminology that task-specific ASR systems may lack. Motivated by these advantages, various studies have explored the integration of LLMs into ASR \citep{chen2024hyporadise,pu2023multi}, Machine Translation (MT) \citep{koneru2023contextual}, and ST \citep{hu2024gentranslate}.

\begin{figure*}[!ht]
\includegraphics[width=\textwidth]{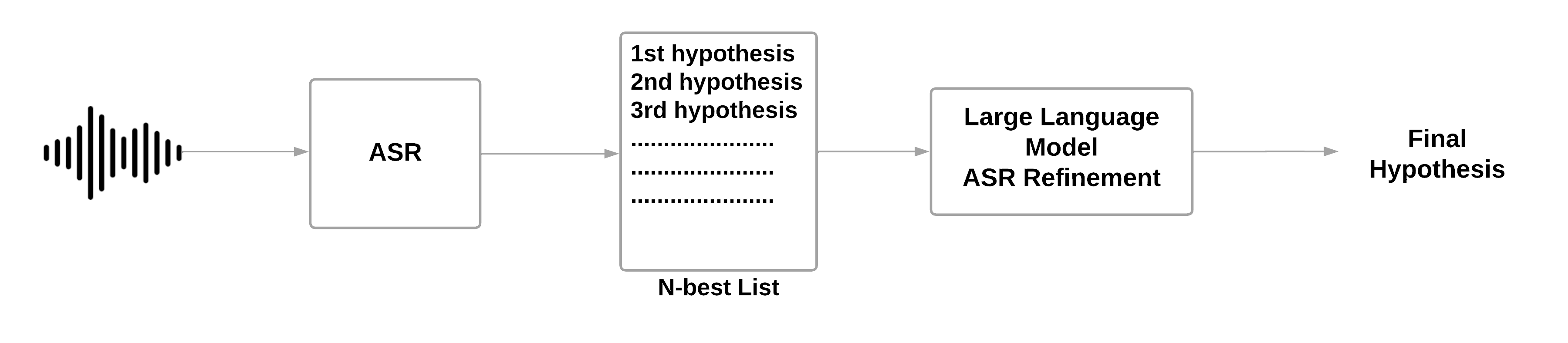}
 \caption{\textbf{ASR Refinement}: The ASR system generates a few candidate hypotheses with beam search, and the LLM generates a new hypothesis based on all the candidates as proposed in \citet{chen2024hyporadise}. We use the top 5 candidates in all our experiments.}
\label{fig:asrpe}
\end{figure*}

\citet{chen2024hyporadise} employ the LLM to generate a new hypothesis based on the N-best list of the ASR model. This strategy relies on the observation that N-best lists tend to exhibit enough diversity, especially during uncertain conditions, allowing accurate transcript prediction by examining the list. On the other hand, for MT, \citet{koneru2023contextual} proposes leveraging the LLM to automatically postedit translations by analyzing the source and hypothesis documents to rectify contextual errors. Both approaches are system-agnostic and have demonstrated successful enhancement of system quality. Furthermore, it is also the case that cascaded systems are shown to be superior than end to end systems in previous IWSLT findings and submissions making the leveraging of LLMs easily compatible. \citep{agarwal2023findings,liu2023kit}.

Our system builds on these two approaches to effectively use the LLMs to improve the cascaded ST pipeline by refining the intermediate outputs at both ASR and MT while maintaining its modular structure. We utilize pre-trained models to create the individual components and fine-tune them with the allowed data. Specifically, we employ WavLM \citep{chen2022wavlm} and MBART50 \citep{liu2020multilingual} to initialize the ASR, and NLLB-200 (3.3B) \citep{costa2022no} for the MT module. As for the LLM, we opt for Mistral 7B Instruction-Tuned \citep{jiang2023mistral}, considering it to be the most recent model within the allowable options.

We present our main findings below:

\begin{itemize}
    \item We demonstrate that LLMs can be tailored to enhance both ASR (Section \ref{asrresults}) and MT systems (Section \ref{stresults}), resulting in an absolute improvement of $0.3\%$ in Word Error Rate and $0.65\%$ in COMET, respectively, on the tst2019 test set.

    \item While we observe significant enhancements in in-domain scenarios, we find that these techniques are not applicable in challenging scenarios (such as Overlapping Speakers, Background noise, etc.) due to poor ASR performance.

    \item We demonstrate that employing chunked long-form decoding\footnote{We derive the terminology from this \href{https://huggingface.co/blog/asr-chunking}{blog post}.} significantly improves ASR performance in challenging scenarios, such as the case of the ITV dev set. Specifically, we observe a decrease in the word error rate from $37.83\%$ to $30.98\%$
\end{itemize}

\section{Data}
\label{data}
This section describes the evaluation and training data we use in our experiments. For evaluation, we report results on the  tst2019 and ACLdev \citep{salesky-2023-evaluating} test sets to compare with findings from previous works \citep{anastasopoulos-etal-2021-findings, agarwal2023findings}. We also use the EPTV (European Parlament activities), Itv (TV Series), and Peloton (Fitness TV) dev sets from the subtitling track consisting of overlapping speakers with different accents to evaluate the ASR performance in challenging scenarios.

As the data conditions did not change from IWSLT23 to this year, we rely on the data processed from last year's submission (KIT'23) \citep{liu2023kit}. For the training data of ASR, we use the same system that used Common Voice \citep{cv}, LibriSpeech \citep{librispeech}, MuST-C v2 \citep{di-gangi-etal-2019-must}, TED-LIUM v3 \citep{tedlium3}, and VoxPopuli \citep{wang-etal-2021-voxpopuli}.

While for MT fine-tuning, we use the cleaned training data from last year created from the available parallel data. This includes Europarl v7 and v10 \citep{koehn-2005-europarl}, NewsCommentary v16, OpenSubtitles v2018 \citep{lison-tiedemann-2016-opensubtitles2016}, Tatoeba \citep{tiedemann-2012-parallel}, ELRC-CORDIS\_News and TED2020 \citep{reimers-gurevych-2020-making} and consists in total of 23 million sentence pairs. For the rest of the paper, we refer to the full parallel data as \textit{seed} and TED2020 as \textit{in-domain}.

\section{Overview}
In this section, we provide an overview of our proposed cascaded system, detailing each individual component. First, the input audio is sent to the ASR system, which undergoes segmentation, and N-best lists are generated for each segmented utterance. Next, the top candidates in the N-best list are fed as input to the LLM, which is trained to refine the ASR output and generate a final ASR hypothesis. Following this, the final ASR hypotheses are passed on to the sentence-level MT system, which produces translations. Finally, the sentence-level automatic transcripts and translations are fed into another adapted LLM, which automatically post-edits and generates a coherent document translation of the talk.

\subsection{Automatic Speech Recognition}
\label{asr}

We employed the ASR model from our previous year's submission \citep{liu2023kit}, considering its effectiveness in transcribing the TED domain. For initialization, we utilized WavLM and mBART50 for the encoder and decoder, respectively, before fine-tuning on the ASR data described in Section \ref{data}. However, we encountered below-par ASR performance on the challenging sets EPTV, Itv, and Peloton.


 We identified several issues that hindered the effectiveness of our ASR model with these sets. Firstly, the model itself was trained on single-talker datasets but inferred with multi-talker noisy datasets, leading to a mismatch in data distribution. Secondly, our typical use of the SHAS model for audio segmentation introduced challenges, as it sometimes missed segmentations and overlooked segments containing human speech. 
 



Data shift is difficult to handle when the training dataset has not changed since last year. We focused more on handling the latter by incorporating long-form decoding. The key idea is to better use context (at the text or signal level) for decoding. The long audio file is chunked into smaller segments with a small overlap between adjacent segments. The model is run over each chunk, and the inferred text is joined at the strides by finding the longest common sequence between overlaps.

\subsection{ASR refinement}
\label{asrrefine}

\begin{figure*}[!ht]
\includegraphics[width=\textwidth]{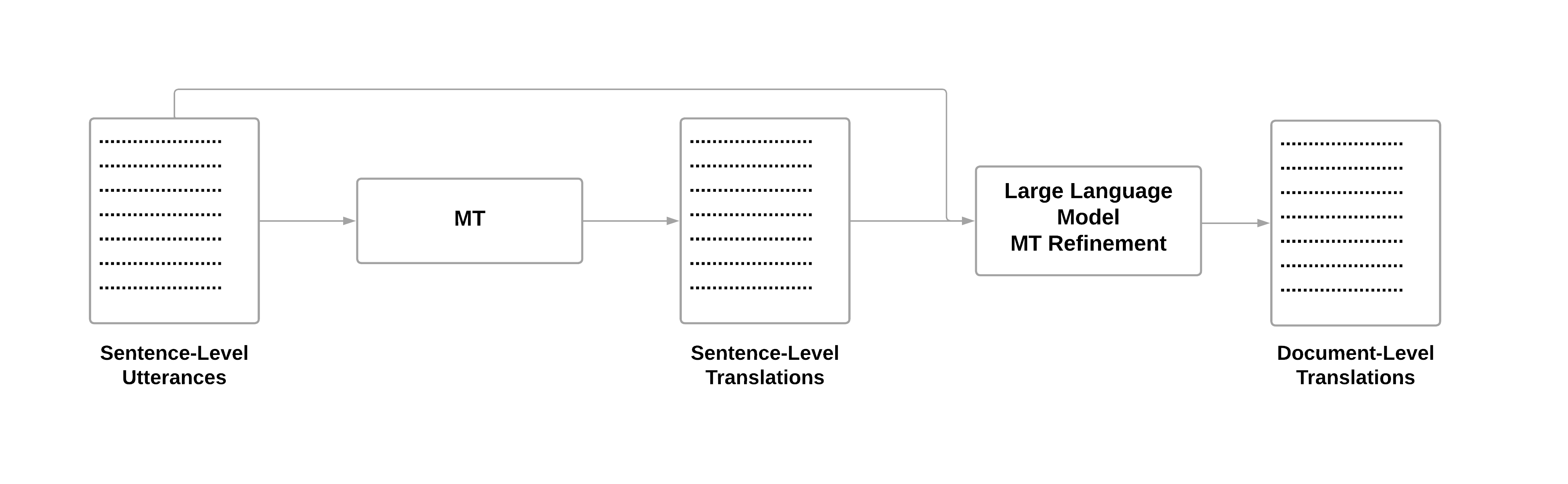}
 \caption{\textbf{Document Level MT Refinement}: The LLM trained to post-edit uses sentence-level transcripts and translations to generate a final document-level coherent and consistent translation.}
\label{fig:mtpe}
\end{figure*}

Once we have generated the N-best list, we select the top 5 candidates and utilize an LLM to produce the final hypothesis as shown in Figure \ref{fig:asrpe}. In this step, we can adapt the LLM to the task using either few-shot prompting or LoRA fine-tuning techniques. We choose to fine-tune the LLM with adapters based on the findings from \citep{chen2024hyporadise}. However, it is crucial to train the LLM under conditions that simulate the test environment, where it should fix errors of our ASR output rather than on the whisper generated in \citet{chen2024hyporadise}.

To generate the dataset for fine-tuning, we perform inference on our in-domain training data using the gold segmentation. We create pairs comprising the N-best list and the corresponding reference. It is worth noting that we utilized the same data to train the ASR system, which is not ideal. However, resource constraints prevented us from following the augmentation procedure that mitigates this, which we explain further in Section \ref{docpe}. Despite this limitation, manual analysis revealed that the ASR did not memorize the training data and produced similar N-best lists to those observed in the test conditions.

Following this, we fine-tuned the Mistral 7B Instruction-tuned LLM \citep{jiang2023mistral} using QLoRA \citep{dettmers2024qlora}, to predict the gold reference based on the top candidates (see the prompt format below). Importantly, we chose not to shuffle the order of the top candidates when providing it in the prompt, as doing so would eliminate the ranking information provided to the LLM, which could be crucial for its performance.

\begin{verbatim}
Punctuate and Post-edit the hypothesis
based on the predictions:
Hyp 1 <SS> Hyp 2 <SS> Hyp 3 ..
Post-edited Hypothesis:
Gold Reference 
\end{verbatim}

\subsection{Machine Translation}
\label{mt}

For building the MT system, we leverage the strong pre-trained model NLLB 200 3.3B \citep{costa2022no} that is allowed in the constrained plus LLM track. We perform a two-step fine-tuning approach. Initially, we fine-tune the model on the \textit{seed} data to adapt it to the spoken language domain. Subsequently, in the second step, we conduct in-domain fine-tuning on TED (in-domain) data, given its significance as one of the primary test sets in the offline track. Additionally, we implement checkpoint averaging to improve generalization with the last 3 checkpoints.

\subsubsection{Restoring Punctuations}

It is important to note that the ASR outputs lack punctuation. Therefore, we conducted experiments with two punctuators. First, we utilized the punctuations generated from the LLM ASR refinement process described in Section \ref{asrrefine}. Second, we employed a DeltaLM-based punctuation model, which was utilized in our previous year's submission \citep{liu2023kit}. We observed that while the punctuations generated by the LLM were semantically correct, they often resulted in long sequences and led to a degradation in MT performance. As a result, we decided to opt for the second choice and segment the text into sentences using manually crafted rules.

\begin{table*}[]
\centering
\begin{tabular}{@{}lll@{}}
\toprule
                       & tst2019 & ACLdev2023 \\ \midrule
KIT'23 \citep{liu2023kit} ASR              & 3.1     & 11.3       \\
KIT'23 ASR + LLM Refine & 2.8     & 10.6       \\ \bottomrule
\end{tabular}
\caption{ASR word error rate scores on tst2019 and ACLdev2023 test sets. $+$ LLM refine indicates that the N-best list was post-edited to generate the final hypothesis.}
\label{tab:asr_llm}
\end{table*}

\subsection{Document-level Automatic Post-Editing}
\label{docpe}

After translating the individual sentences with the fine-tuned NLLB, the outputs are not coherent as they are translated in isolation. Moreover, any ASR errors that might be fixed by observing the full document will be translated incorrectly. To mitigate this, we perform an additional step of document-level automatic post-editing using the source transcripts and sentence translations shown in Figure \ref{fig:mtpe}.

Similar to the situation outlined in Section \ref{asrrefine}, we encountered a lack of data for fine-tuning the LLM for document-level post-editing. Hence, we adopted the approach proposed by \citet{koneru2023contextual} to create the dataset. We divided the in-domain TED data into two halves, each containing English audio, English transcript, and German translation. Subsequently, we fine-tuned MT models on each half using the pre-trained models described in Sections \ref{asr} and \ref{mt}. Following this, we conducted inference using the gold segmentation with our ASR and MT models trained on one half to the other half. This procedure generated a synthetic dataset with noisy ASR input, MT predictions, and corresponding gold references, leveraging the provided segmentation in the data. 

We then use the synthetic dataset to create instances of document-level post-editing. We go through each talk and divide the transcripts into chunks, each chunk containing a maximum of 256 tokens corresponding to the LLM tokenizer. Then for each chunk, we use the transcript, hypothesis and reference to transform them into the format below and train the LLM to predict the gold reference given the noisy transcript and sentence-level hypothesis.

\begin{verbatim}
Noisy English Transcript: 
ASR Hyp 1 <SS> ASR Hyp 2 <SS> ....
German Translations:
MT Hyp 1 <SS> MT Hyp 2 <SS> ....
Post-Edited German Translations:
Ref 1 <SS> Ref 2 <SS> ....
\end{verbatim}


We use the delimiter "<SS>" to align with the input and perform sentence-level evaluation. Then, we again fine-tune the Mistral 7B Instruction-tuned LLM \citep{jiang2023mistral} using QLoRA \citep{dettmers2024qlora}, training it to predict the gold reference given the noisy transcript and translations. We employ the sliding window with payload strategy during decoding as described in \citet{koneru2023contextual}.

\section{Results}

\subsection{Automatic Speech Recognition}
\label{asrresults}

To evaluate the benefit of the additional ASR refinement step described in Section \ref{asrrefine}, we compare the word error rate of our ASR system before and after post-editing, as shown in Table \ref{tab:asr_llm}. The ASR performance improves in both cases, with a higher absolute improvement observed in the ACLdev set. The LLM is particularly beneficial in the ACLdev set, given that it contains terminology from the scientific domain where the LLM excels. We also observe a relative improvement of $10\%$ in the TED talks, indicating that ASR refinement is beneficial.

\begin{table*}[]
\centering
\begin{tabular}{@{}cccc@{}}
\toprule
\textit{Model}       & \textit{EPTV} & \textit{ITV} & \textit{Peloton} \\ \midrule
KIT'23 ASR             & 26.43         & 37.83        & \textbf{18.93}            \\
KIT'23 ASR  + Gold Seg & \textbf{16.84}         & 37.21        & 25.88            \\
KIT'23 ASR  + long-form & 17.54         & \textbf{30.98}        & 20.79   \\
Seamless v2 \citep{barrault2023seamlessm4t}         & 40.94         & 56.94        & 43.47            \\ \bottomrule
\end{tabular}
\caption{ASR word error rate scores on the EPTV, Peloton and ITV dev set. Best scores for each set are highlighted in bold.}
\label{tab:asr_eptv}
\end{table*}

However, the performance of the same ASR system on the challenge set was below par. We conducted additional ablation studies and present the results in Table \ref{tab:asr_eptv} for the challenge dev sets. We compared last year's ASR system with three conditions: providing gold segmentation, utilizing long-form decoding, and using the recently developed Seamless V2 \citep{barrault2023seamlessm4t}.

We observed that providing gold segmentation achieved a score of $16.84$, demonstrating its crucial role in handling this challenging set for EPTV. Moreover, long-form decoding significantly narrowed the gap, decreasing the word error rate for both EPTV and ITV. Meanwhile, our ASR shows the best performance for Peloton without any modifications. Additionally, we evaluated Seamless to assess its robustness and found that its performance was severely lacking in comparison.

Based on these results, we use the ASR with standard segmentation for \textit{TED} and \textit{Peloton} test sets. For EPTV and ITV, we use the ASR system with long-form decoding. We found that the LLM cannot refine the N-best list given the poor WER of KIT'23 ASR for the latter test sets and generates long sequences with repetitions for most utterances. \textbf{Therefore, we do not perform any ASR or MT LLM refinements for \textit{ITV} and \textit{EPTV} sets and generate translations with a standard cascaded ST pipeline.}

\subsection{Cascaded Speech Translation}
\label{stresults}

In this section, we evaluate the final quality of our cascaded ST using the mwerSegmenter to realign the hypothesis with the reference segmentation. We report results with BLEU \citep{papineni-etal-2002-bleu} and Chrf2 \citep{popovic-2015-chrf} computed by Sacrebleu \citep{post-2018-call}. We also report the COMET \citep{rei-etal-2022-comet} score using the default model\footnote{Unbabel/wmt22-comet-da}.

\subsection{Two-step Fine-tuning}

We presented a two-step fine-tuning approach to adapt our MT system in Section \ref{mt} to the target domain. We report the translation quality on tst2019 test set with this approach (last row) and other models for comparison in Table \ref{tab:st_base}.

\begin{table}[]
\resizebox{\columnwidth}{!}{
\begin{tabular}{@{}cccc@{}}
\toprule
\textit{Model}        & \multicolumn{3}{c}{\textit{tst2019}}  \\ \midrule
                      & BLEU & Chrf2         & COMET          \\ \midrule
KIT'23 TED*    & \textbf{28.4} & \textbf{58.8} & \textbf{78.87} \\
NLLB 3.3B              & 26.6 & 57.7          & 77.41          \\
Seamless v2             & 25.5 & 57.0            & 76.65 \\
NLLB 3.3B + Seed       & 26.9 & 57.9          & 77.87          \\
NLLB 3.3B + Seed + TED & 27.6 & 58.5          & 78.49          \\ \bottomrule
\end{tabular}
}
\caption{MT scores using KIT'23 ASR as input calculated by resegmenting with mwerSegmenter. * indicates an unconstrained system that was trained on the same data sources but in more languages than what is allowed for IWSLT24. TED indicates the model adapted for TED and not ACLdev which was the official submission from KIT for IWSLT23}
\label{tab:st_base}
\end{table}

\begin{table*}[!t]
\centering
\resizebox{2\columnwidth}{!}{
\begin{tabular}{@{}c|ccc|ccc@{}}
\toprule
\multirow{2}{*}{\textit{Model}}               & \multicolumn{3}{c|}{\textit{tst2019}}          & \multicolumn{3}{c}{ACLdev2023}               \\ \cmidrule(l){2-7} 
                                              & BLEU          & Chrf2         & COMET          & BLEU          & Chrf2       & COMET          \\ \midrule
NLLB 3.3                                      & 26.6          & 57.7          & 77.41          & 35.0            & 63.9        & 74.83          \\
NLLB 3.3 Seed                                 & 26.9          & 57.9          & 77.87          & 34.7          & 63.8        & 75.67          \\
ASR Refine + NLLB 3.3 Seed                    & 27.3          & 58.3          & 78.32          & 36.1          & \textbf{65.0} & 77.59          \\
ASR Refine + NLLB 3.3 Seed + TED              & 28.3          & 58.8          & 78.98          & 34.8          & 63.7        & 77.25          \\
ASR Refine + NLLB 3.3 Seed + TED  + Doc APE & \textbf{28.7} & \textbf{59.1} & \textbf{79.63} & \textbf{36.4} & 64.5        & \textbf{78.64} \\ \bottomrule
\end{tabular}
}
\caption{MT scores using KIT'23 ASR as input calculated by resegmenting with mwerSegmenter. ASR Refine indicates an additional ASR refinement step with the LLM. Seed and TED indicate fine-tuning the NLLB 3.3 with seed alone or a two-step process with additional fine-tuning on TED. Doc APE indicates an LLM post-editing refinement to generate a coherent and consistent document. Best scores in each metric and test set are highlighted in bold.}
\label{tab:st_full}
\end{table*}

Firstly, we observe that Seamless performs inferiorly to NLLB across all translation metrics. Consequently, we proceeded with NLLB for further experiments. 

Subsequently, fine-tuning the seed parallel data improved quality across all metrics, notably increasing the score from $77.41$ to $77.87$ in COMET. Following this, with the assistance of second-step fine-tuning, we observed further improvements, resulting in scores reaching $78.49$. However, it is important to note that this system still lags behind last year's submission, which was specifically adapted to the TED domain. Nevertheless, it's worth highlighting that this system was trained across multiple languages, placing it in the unconstrained condition for IWSLT24. Moreover, we could not replicate a similar adaptation process for NLLB due to resource and time constraints.

\subsection{LLM Refinement}

We proposed improving the ASR outputs and converting sentence-level to document-level translations using fine-tuned LLMs. We evaluate the benefits of the individual steps and report the results of our final cascaded ST system in Table \ref{tab:st_full} on tst2019 and ACLdev2023 test sets.

First, the benefits of two-step fine-tuning, ASR refinement, and document post-editing complement each other. Using KITs 23 ASR with NLLB 3.3 B as a baseline, we obtained $77.41$ COMET in tst2019 test set. However, including all enhancements led to a total improvement of $2.23$ COMET points. Furthermore, the improvements are consistent with both lexical and neural metrics.

Next, we observed that integrating LLMs provides significant benefits in the ACLdev set compared to the TED dev sets. This is plausible due to scientific terminology and accented speakers in the ACLdev set. Both of these challenges are well-suited for LLMs, as the quality of the initial systems is sufficient to utilize context and rectify mistakes reliably.

\section{Conclusion}
This system paper presented KIT's submission for the offline track in the constrained + LLMs condition, focusing on the English-to-German translation direction. Using modular techniques, we successfully integrated LLMs into any cascaded ST pipeline. Additionally, we highlighted the benefits of long-form decoding in scenarios involving noisy and overlapping speech. 

For future work, we aim to explore robust techniques for integrating LLMs that can effectively handle challenging scenarios where ASR quality is sub-par. Furthermore, the translation's latency is quite high as it needs to call the LLM twice. However, integrating quality estimation techniques to decide when we need the LLM can limit the effects of the high latency problem.

\section*{Acknowledgments}
This work is partly supported by the Helmholtz Programme-oriented Funding, with project number 46.24.01, named AI for Language Technologies, funding from the pilot program Core-Informatics of the Helmholtz Association (HGF).
It also received partial support from the European Union’s Horizon research and innovation programme under grant agreement No 101135798, project Meetween (My Personal AI Mediator for Virtual MEETtings BetWEEN People). The work was partly performed on the HoreKa supercomputer funded by the Ministry of Science, Research and the Arts Baden-Württemberg and by the Federal Ministry of Education and Research.

\bibliography{anthology,main}

\appendix

\section{Appendix}
\label{sec:appendix}


We use the transformers library \citep{wolf2019huggingface} for fine-tuning our ASR and LLM and the fairseq toolkit \citep{ott2019fairseq} for fine-tuning NLLB 3.3B. For the ASR training, we set the \textit{batch size} to $384$, resulting in approximately 128 minutes per batch. We employ a \textit{warmup strategy} over $2,000$ steps and a total of $100,000$ training steps. The \textit{learning rate} is initialized to $1e-4$.

For the NLLB fine-tuning experiments, we use a \textit{learning rate} set to $5e-5$, \textit{label smoothing} to $0.1$, \textit{drop out} to $0.1$, \textit{attention drop-out} to $0.1$. We use the \textit{Adam optimizer} with \textit{betas} to $(0,9,0.98)$ and the remaining optimizer parameters to default. We used a \textit{batch size} of maximum $3096$ making one step, \textit{update-freq} to $16$ and validating on the dev set after every epoch. We stopped the training after the dev loss did not increase after 10 epochs.  
For fine-tuning the LLM with QLoRA we use the peft \citep{peft} along with the transformers library. We add LoRA adapters to the target modules [\textit{q\_proj, k\_proj, v\_proj, o\_proj, gate\_proj, up\_proj, down\_proj}]. We set the \textit{adapter rank} to $16$, \textit{alpha} to $32$ and \textit{lora dropout} to $0.1$.  We use a \textit{batch size of $8$}, \textit{learning rate} of $5e-5$ with other parameters set to default. After every 200 steps, we validate and terminate the training if it does not improve 10 consecutive times.

During inference, we use beam search for all ASR, MT and LLM components. The ASR and MT decode with \textit{beam size} of 5, whereas the LLM does it with \textit{beam size} of $3$.

\end{document}